# GraphLab: A New Framework For Parallel Machine Learning

**Yucheng Low**
Carnegie Mellon University
ylow@cs.cmu.edu

**Joseph Gonzalez**
Carnegie Mellon University
jegonzal@cs.cmu.edu

**Aapo Kyrola**
Carnegie Mellon University
akyrola@cs.cmu.edu

**Danny Bickson**
Carnegie Mellon University
bickson@cs.cmu.edu

**Carlos Guestrin**
Carnegie Mellon University
guestrin@cs.cmu.edu

**Joseph Hellerstein**
UC Berkeley
hellerstein@cs.berkeley.edu

## Abstract

Designing and implementing *efficient*, *provably correct* parallel machine learning (ML) algorithms is challenging. Existing high-level parallel abstractions like MapReduce are insufficiently expressive while low-level tools like MPI and Pthreads leave ML experts repeatedly solving the same design challenges. By targeting common patterns in ML, we developed GraphLab, which improves upon abstractions like MapReduce by compactly expressing asynchronous iterative algorithms with sparse computational dependencies while ensuring data consistency and achieving a high degree of parallel performance. We demonstrate the expressiveness of the GraphLab framework by designing and implementing parallel versions of belief propagation, Gibbs sampling, Co-EM, Lasso and Compressed Sensing. We show that using GraphLab we can achieve excellent parallel performance on large scale real-world problems.

## 1 INTRODUCTION

Exponential gains in hardware technology have enabled sophisticated machine learning (ML) techniques to be applied to increasingly challenging real-world problems. However, recent developments in computer architecture have shifted the focus away from frequency scaling and towards parallel scaling, threatening the future of sequential ML algorithms. In order to benefit from future trends in processor technology and to be able to apply rich structured models to rapidly scaling real-world problems, the ML community must directly confront the challenges of parallelism.

However, designing and implementing *efficient* and *provably correct* parallel algorithms is extremely challenging. While low level abstractions like MPI and Pthreads provide powerful, expressive primitives, they force the user to address hardware issues and the challenges of parallel data representation. Consequently, many ML experts have turned to high-level abstractions, which dramatically simplify the design and implementation of a *restricted* class of parallel algorithms. For example, the MapReduce abstraction [Dean and Ghemawat, 2004] has been successfully applied to a broad range of ML applications [Chu et al., 2006, Wolfe et al., 2008, Panda et al., 2009, Ye et al., 2009].

However, by restricting our focus to ML algorithms that are naturally expressed in MapReduce, we are often forced to make overly simplifying assumptions. Alternatively, by coercing efficient sequential ML algorithms to satisfy the restrictions imposed by MapReduce, we often produce inefficient parallel algorithms that require many processors to be competitive with comparable sequential methods.

In this paper we propose GraphLab, a new parallel framework for ML which exploits the *sparse structure* and common *computational patterns* of ML algorithms. GraphLab enables ML experts to easily design and implement efficient scalable parallel algorithms by composing problem specific computation, data-dependencies, and scheduling. We provide an efficient *shared-memory* implementation[1] of GraphLab and use it to build parallel versions of four popular ML algorithms. We focus on the shared-memory multiprocessor setting because it is both ubiquitous and has few effective high-level abstractions. We evaluate the algorithms on a 16-processor system and demonstrate state-of-the-art performance. *Our main contributions include*:

- A graph-based data model which simultaneously represents data and computational dependencies.
- A set of concurrent access models which provide a range of sequential-consistency guarantees.
- A sophisticated modular scheduling mechanism.
- An aggregation framework to manage global state.
- GraphLab implementations and experimental evaluations of parameter learning and inference in graphical models, Gibbs sampling, CoEM, Lasso and compressed sensing on real-world problems.

---

[1] The C++ reference implementation of the GraphLab is available at http://select.cs.cmu.edu/code.

## 2 EXISTING FRAMEWORKS

There are several existing frameworks for designing and implementing parallel ML algorithms. Because GraphLab generalizes these ideas and addresses several of their critical limitations we briefly review these frameworks.

### 2.1 MAP-REDUCE ABSTRACTION

A program implemented in the MapReduce framework consists of a `Map` operation and a `Reduce` operation. The `Map` operation is a function which is applied independently and in parallel to each datum (e.g., webpage) in a large data set (e.g., computing the word-count). The `Reduce` operation is an aggregation function which combines the `Map` outputs (e.g., computing the total word count). MapReduce performs optimally only when the algorithm is *embarrassingly parallel* and can be decomposed into a large number of independent computations. The MapReduce framework expresses the class of ML algorithms which fit the Statistical-Query model [Chu et al., 2006] as well as problems where feature extraction dominates the run-time.

The MapReduce abstraction fails when there are *computational dependencies* in the data. For example, MapReduce can be used to extract features from a massive collection of images but cannot represent computation that depends on small overlapping subsets of images. This critical limitation makes it difficult to represent algorithms that operate on structured models. As a consequence, when confronted with large scale problems, we often abandon rich structured models in favor of overly simplistic methods that are amenable to the MapReduce abstraction.

Many ML algorithms *iteratively* transform parameters during both learning and inference. For example, algorithms like Belief Propagation (BP), EM, gradient descent, and even Gibbs sampling, iteratively refine a set of parameters until some termination condition is achieved. While the MapReduce abstraction can be invoked iteratively, it does not provide a mechanism to directly encode iterative computation. As a consequence, it is not possible to express sophisticated scheduling, automatically assess termination, or even leverage basic data persistence.

The popular implementations of the MapReduce abstraction are targeted at large data-center applications and therefore optimized to address node-failure and disk-centric parallelism. The overhead associated with the fault-tolerant, disk-centric approach is unnecessarily costly when applied to the typical cluster and multi-core settings encountered in ML research. Nonetheless, MapReduce is used in small clusters and even multi-core settings [Chu et al., 2006]. The GraphLab implementation[2] described in this paper does not address fault-tolerance or parallel disk access and instead

---

[2]The GraphLab abstraction is intended for both the multicore and cluster settings and a distributed, fault-tolerant implementation is ongoing research.

---

assumes that processors do not fail and all data is stored in shared-memory. As a consequence, GraphLab does not incur the unnecessary disk overhead associated with MapReduce in the multi-core setting.

### 2.2 DAG ABSTRACTION

In the DAG abstraction, parallel computation is represented as a directed acyclic graph with data flowing along edges between vertices. Vertices correspond to functions which receive information on inbound edges and output results to outbound edges. Implementations of this abstraction include Dryad [Isard et al., 2007] and Pig Latin [Olston et al., 2008].

While the DAG abstraction permits rich computational dependencies it does not naturally express iterative algorithms since the structure of the dataflow graph depends on the number of iterations (which must therefore be known prior to running the program). The DAG abstraction also cannot express dynamically prioritized computation.

### 2.3 SYSTOLIC ABSTRACTION

The Systolic abstraction [Kung and Leiserson, 1980] (and the closely related Dataflow abstraction) extends the DAG framework to the iterative setting. Just as in the DAG Abstraction, the Systolic abstraction forces the computation to be decomposed into small atomic components with limited communication between the components. The Systolic abstraction uses a directed graph $G = (V, E)$ which is not necessarily acyclic) where each vertex represents a processor, and each edge represents a communication link. In a single iteration, each processor reads all incoming messages from the in-edges, performs some computation, and writes messages to the out-edges. A barrier synchronization is performed between each iteration, ensuring all processors compute and communicate in lockstep.

While the Systolic framework can express iterative computation, it is unable to express the wide variety of update schedules used in ML algorithms. For example, while gradient descent may be run within the Systolic abstraction using a **Jacobi schedule** it is not possible to implement coordinate descent which requires the more sequential **Gauss-Seidel schedule**. The Systolic abstraction also cannot express the dynamic and specialized structured schedules which were shown by Gonzalez et al. [2009a,b] to dramatically improve the performance of algorithms like BP.

## 3 THE GRAPHLAB ABSTRACTION

By targeting common patterns in ML, like sparse data dependencies and asynchronous iterative computation, GraphLab achieves a balance between low-level and high-level abstractions. Unlike many low-level abstractions (e.g., MPI, PThreads), GraphLab insulates users from the complexities of synchronization, data races and deadlocks by providing a high level data representation through the **data graph** and automatically maintained data-consistency guarantees through configurable **consistency models**. Unlike many high-level abstractions (i.e., MapReduce), GraphLab can express complex computational dependencies using the **data graph** and provides sophisticated **scheduling primitives** which can express iterative parallel algorithms with dynamic scheduling.

To aid in the presentation of the GraphLab framework we use Loopy Belief Propagation (BP) [Pearl, 1988] on pairwise Markov Random Fields (MRF) as a running example. A pairwise MRF is an undirected graph over random variables where edges represent interactions between variables. Loopy BP is an approximate inference algorithm which estimates the marginal distributions by iteratively recomputing parameters (messages) associated with each edge until some convergence condition is achieved.

## 3.1 DATA MODEL

The GraphLab data model consists of two parts: a directed **data graph** and a **shared data table**. The data graph $G = (V, E)$ encodes both the problem specific *sparse computational structure* and directly modifiable program state. The user can associate arbitrary blocks of data (or parameters) with each vertex and directed edge in $G$. We denote the data associated with vertex $v$ by $D_v$, and the data associated with edge $(u \to v)$ by $D_{u \to v}$. In addition, we use $(u \to *)$ to represent the set of all outbound edges from $u$ and $(* \to v)$ for inbound edges at $v$. To support globally shared state, GraphLab provides a **shared data table** (SDT) which is an associative map, $\mathbf{T}[\text{Key}] \to \text{Value}$, between keys and arbitrary blocks of data.

In the Loopy BP, the data graph is the pairwise MRF, with the vertex data $D_v$ to storing the node potentials and the directed edge data $D_{u \to v}$ storing the BP messages. If the MRF is sparse then the data graph is also sparse and GraphLab will achieve a high degree of parallelism. The SDT can be used to store shared hyper-parameters and the global convergence progress.

## 3.2 USER DEFINED COMPUTATION

Computation in GraphLab can be performed either through an **update function** which defines the local computation, or through the **sync mechanism** which defines global aggregation. The Update Function is analogous to the `Map` in MapReduce, but unlike in MapReduce, update unctions are permitted to access and modify *overlapping* contexts in the graph. The sync mechanism is analogous to the `Reduce` operation, but unlike in MapReduce, the sync mechanism runs concurrently with the update functions.

### 3.2.1 Update Functions

A GraphLab **update function** is a stateless user-defined function which operates on the data associated with small neighborhoods in the graph and represents the core element

**Algorithm 1**: Sync Algorithm on $k$

$t \leftarrow r_k^{(0)}$
**foreach** $v \in V$ **do**
$\quad t \leftarrow \text{Fold}_k(D_v, t)$
$\mathbf{T}[k] \leftarrow \text{Apply}_k(t)$

of computation. For every vertex $v$, we define $\mathcal{S}_v$ as the neighborhood of $v$ which consists of $v$, its adjacent edges (both inbound and outbound) and its neighboring vertices as shown in Fig. 1(a). We define $D_{\mathcal{S}_v}$ as the data corresponding to the neighborhood $\mathcal{S}_v$. In addition to $D_{\mathcal{S}_v}$, update functions also have *read-only* access, to the shared data table $\mathbf{T}$. We define the application of the update function $f$ to the vertex $v$ as the state mutating computation:

$$D_{\mathcal{S}_v} \leftarrow f(D_{\mathcal{S}_v}, \mathbf{T}).$$

We refer to the neighborhood $\mathcal{S}_v$ as the **scope** of $v$ because $\mathcal{S}_v$ defines the extent of the graph that can be accessed by $f$ when applied to $v$. For notational simplicity, we denote $f(D_{\mathcal{S}_v}, \mathbf{T})$ as $f(v)$. A GraphLab program may consist of multiple update functions and it is up to the scheduling model (see Sec. 3.4) to determine which update functions are applied to which vertices and in which parallel order.

### 3.2.2 Sync Mechanism

The **sync mechanism** aggregates data across all vertices in the graph in a manner analogous to the Fold and Reduce operations in functional programming. The result of the sync operation is associated with a particular entry in the Shared Data Table (SDT). The user provides a key $k$, a **fold function** (Eq. (3.1)), an **apply function** (Eq. (3.3)) as well as an initial value $r_k^{(0)}$ to the SDT and an optional **merge function** used to construct parallel tree reductions.

$$r_k^{(i+1)} \leftarrow \text{Fold}_k\left(D_v, r_k^{(i)}\right) \quad (3.1)$$

$$r_k^l \leftarrow \text{Merge}_k\left(r_k^i, r_k^j\right) \quad (3.2)$$

$$\mathbf{T}[k] \leftarrow \text{Apply}_k(r_k^{(|V|)}) \quad (3.3)$$

When the sync mechanism is invoked, the algorithm in Alg. 1 uses the $\text{Fold}_k$ function to *sequentially* aggregate data across all vertices. The $\text{Fold}_k$ function obeys the same consistency rules (described in Sec. 3.3) as update functions and is therefore able to *modify* the vertex data. If the $\text{Merge}_k$ function is provided a *parallel* tree reduction is used to combine the results of multiple parallel folds. The $\text{Apply}_k$ then finalizes the resulting value (e.g., rescaling) before it is written back to the SDT with key $k$.

The sync mechanism can be set to run periodically in the background while the GraphLab engine is actively applying update functions or on demand triggered by update functions or user code. If the sync mechanism is executed in the background, the resulting aggregated value may not be globally consistent. Nonetheless, many ML applications are robust to approximate global statistics.

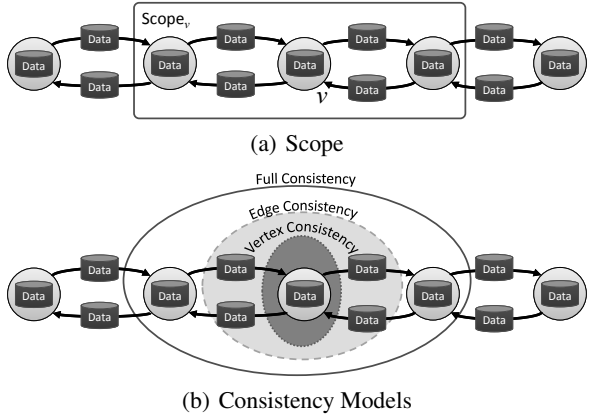

(a) Scope

(b) Consistency Models

Figure 1: **(a)** The scope, $\mathcal{S}_v$, of vertex $v$ consists of all the data at the vertex $v$, its inbound and outbound edges, and its neighboring vertices. The update function $f$ when applied to the vertex $v$ can read and modify any data within $\mathcal{S}_v$. **(b)**. We illustrate the 3 data consistency models by drawing their exclusion sets as a ring where no two update functions may be executed simultaneously if their exclusions sets (rings) overlap.

In the context of the Loopy BP example, the update function is the BP message update in which each vertex recomputes its outbound messages by integrating the inbound messages. The sync mechanism is used to monitor the global convergence criterion (for instance, average change or residual in the beliefs). The $\text{Fold}_k$ function accumulates the residual at the vertex, and the $\text{Apply}_k$ function divides the final answer by the number of vertices. To monitor progress, we let GraphLab run the sync mechanism as a periodic background process.

### 3.3 DATA CONSISTENCY

Since scopes may overlap, the simultaneous execution of two update functions can lead to race-conditions resulting in data inconsistency and even corruption. For example, two function applications to neighboring vertices could simultaneously try to modify data on a shared edge resulting in a corrupted value. Alternatively, a function trying to normalize the parameters on a set of edges may compute the sum only to find that the edge values have changed.

GraphLab provides a choice of three data consistency models which enable the user to balance performance and data consistency. The choice of data consistency model determines the extent to which overlapping scopes can be executed simultaneously. We illustrate each of these models in Fig. 1(b) by drawing their corresponding **exclusion sets**. GraphLab guarantees that update functions never simultaneously share overlapping exclusion sets. Therefore larger exclusion sets lead to reduced parallelism by delaying the execution of update functions on nearby vertices.

The **full consistency** model ensures that during the execution of $f(v)$ no other function will read or modify data within $\mathcal{S}_v$. Therefore, parallel execution may only occur on vertices that do not share a common neighbor. The slightly weaker **edge consistency** model ensures that during the execution of $f(v)$ no other function will read or modify any of the data on $v$ or any of the edges adjacent to $v$. Under the edge consistency model, parallel execution may only occur on non-adjacent vertices. Finally, the weakest **vertex consistency** model only ensures that during the execution of $f(v)$ no other function will be applied to $v$. The vertex consistency model is therefore prone to race conditions and should only be used when reads and writes to adjacent data can be done safely (In particular repeated reads may return different results). However, by permitting update functions to be applied simultaneously to neighboring vertices, the vertex consistency model permits maximum parallelism.

Choosing the right consistency model has direct implications to program correctness. One method to prove correctness of a parallel algorithm is to show that it is equivalent to a correct sequential algorithm. To capture the relation between sequential and parallel execution of a program we introduce the concept of **sequential consistency**:

**Definition 3.1** (Sequential Consistency). *A GraphLab program is **sequentially consistent** if for every parallel execution, there exists a sequential execution of update functions that produces an equivalent result.*

The sequential consistency property is typically a sufficient condition to extend algorithmic correctness from the sequential setting to the parallel setting. In particular, if the algorithm is correct under *any* sequential execution of update functions, then the parallel algorithm is also correct if sequential consistency is satisfied.

**Proposition 3.1.** *GraphLab guarantees sequential consistency under the following three conditions:*

1. *The **full consistency** model is used*
2. *The **edge consistency** model is used and update functions do not modify data in adjacent vertices.*
3. *The **vertex consistency** model is used and update functions only access local vertex data.*

In the Loopy BP example the update function only needs to read and modify data on the adjacent edges. Therefore the edge consistency model ensures sequential consistency.

### 3.4 SCHEDULING

The GraphLab **update schedule** describes the order in which update functions are applied to vertices and is represented by a parallel data-structure called the **scheduler**. The **scheduler** abstractly represents a dynamic list of **tasks** (vertex-function pairs) which are to be executed by the GraphLab **engine**.

Because constructing a scheduler requires reasoning about the complexities of parallel algorithm design, the

GraphLab framework provides a collection of base schedules. To represent Jacobi style algorithms (e.g., gradient descent) GraphLab provides a **synchronous scheduler** which ensures that all vertices are updated simultaneously. To represent Gauss-Seidel style algorithms (e.g., Gibbs sampling, coordinate descent), GraphLab provides a **round-robin scheduler** which updates all vertices *sequentially* using the most recently available data.

Many ML algorithms (e.g., Lasso, CoEM, Residual BP) require more control over the tasks that are created and the order in which they are executed. Therefore, GraphLab provides a collection of **task schedulers** which permit update functions to add and reorder tasks. GraphLab provides two classes of task schedulers. The **FIFO** schedulers only permit task creation but do not permit task reordering. The **prioritized** schedules permit task reordering at the cost of increased overhead. For both types of task scheduler GraphLab also provide relaxed versions which increase performance at the expense of reduced control:

|  | Strict Order | Relaxed Order |
|---|---|---|
| **FIFO** | Single Queue | Multi Queue / Partitioned |
| **Prioritized** | Priority Queue | Approx. Priority Queue |

In addition GraphLab provides the **splash scheduler** based on the loopy BP schedule proposed by Gonzalez et al. [2009a] which executes tasks along spanning trees.

In the Loopy BP example, different choices of scheduling leads to different BP algorithms. Using the Synchronous scheduler corresponds to the classical implementation of BP and using priority scheduler corresponds to Residual BP [Elidan et al., 2006].

### 3.4.1 Set Scheduler

Because scheduling is important to parallel algorithm design, GraphLab provides a scheduler construction framework called the **set scheduler** which enables users to *safely* and *easily* compose custom update schedules. To use the set scheduler the user specifies a sequence of vertex set and update function pairs $((S_1, f_1), (S_2, f_2) \cdots (S_k, f_k))$, where $S_i \subseteq V$ and $f_i$ is an update function. This sequence implies the following execution semantics:

**for** $i = 1 \cdots k$ **do**
  Execute $f_i$ on all vertices in $S_i$ in parallel.
  Wait for all updates to complete

The amount of parallelism depends on the size of each set; the procedure is highly sequential if the set sizes are small. Executing the schedule in the manner described above can lead to the majority of the processors waiting for a few processors to complete the current set. However, by leveraging the causal data dependencies encoded in the graph structure we are able to construct an **execution plan** which identifies tasks in future sets that can be executed *early* while still producing an equivalent result.

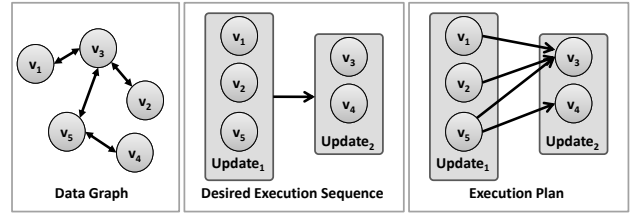

Figure 2: A simple example of the set scheduler planning process. Given the data graph, and a desired sequence of execution where $v_1, v_2$ and $v_5$ are first run in parallel, then followed by $v_3$ and $v_4$. If the edge consistency model is used, we observe that the execution of $v_3$ depends on the state of $v_1, v_2$ and $v_5$, but the $v_4$ only depends on the state of $v_5$. The dependencies are encoded in the execution plan on the right. The resulting plan allows $v_4$ to be immediately executed after $v_5$ without waiting for $v_1$ and $v_2$.

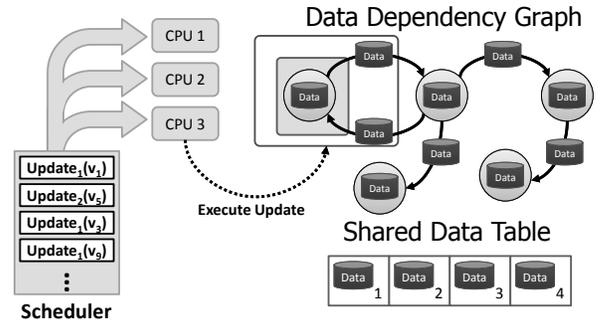

Figure 3: A summary of the GraphLab framework. The user provides a graph representing the computational data dependencies, as well as a SDT containing read only data. The user also picks a scheduling method or defines a custom schedule, which provides a stream of update tasks in the form of (vertex, function) pairs to the processors.

The set scheduler compiles an execution plan by rewriting the execution sequence as a Directed Acyclic Graph (DAG), where each vertex in the DAG represents an update task in the execution sequence and edges represent execution dependencies. Fig. 2 provides an example of this process. The DAG imposes a partial ordering over tasks which can be compiled into a parallel execution schedule using the greedy algorithm described by Graham [1966].

### 3.5 TERMINATION ASSESSMENT

Efficient parallel termination assessment can be challenging. The standard termination conditions used in many iterative ML algorithms require reasoning about the global state. The GraphLab framework provides two methods for termination assessment. The first method relies on the scheduler which signals termination when there are no remaining tasks. This method works for algorithms like Residual BP, which use task schedulers and stop producing new tasks when they converge. The second termination method relies on user provided termination functions which examine the SDT and signal when the algorithm has converged. Algorithms, like parameter learning, which rely on global statistics use this method.

### 3.6 SUMMARY AND IMPLEMENTATION

A GraphLab program is composed of the following parts:

1. A **data graph** which represents the data and computational dependencies.
2. **Update functions** which describe local computation
3. A **Sync mechanism** for aggregating global state.
4. A data **consistency model** (i.e., *Fully Consistent*, *Edge Consistent* or *Vertex Consistent*), which determines the extent to which computation can overlap.
5. **Scheduling primitives** which express the order of computation and may depend dynamically on the data.

We implemented an optimized version of the GraphLab framework in C++ using PThreads. The resulting GraphLab API is available under the LGPL license at http://select.cs.cmu.edu/code. The data consistency models were implemented using race-free and deadlock-free ordered locking protocols. To attain maximum performance we addressed issues related to parallel memory allocation, concurrent random number generation, and cache efficiency. Since mutex collisions can be costly, lock-free data structures and atomic operations were used whenever possible. To achieve the same level of performance for parallel learning system, the ML community would have to repeatedly overcome many of the same *time consuming* systems challenges needed to build GraphLab.

The GraphLab API has the opportunity to be an interface between the ML and systems communities. Parallel ML algorithms built around the GraphLab API automatically benefit from developments in parallel data structures. As new locking protocols and parallel scheduling primitives are incorporated into the GraphLab API, they become immediately available to the ML community. Systems experts can more easily port ML algorithms to new parallel hardware by porting the GraphLab API.

## 4 CASE STUDIES

To demonstrate the expressiveness of the GraphLab abstraction and illustrate the parallel performance gains it provides, we implement four popular ML algorithms and evaluate these algorithms on large real-world problems using a 16-core computer with 4 AMD Opteron 8384 processors and 64GB of RAM.

### 4.1 MRF PARAMETER LEARNING

To demonstrate how the various components of the GraphLab framework can be assembled to build a complete ML "pipeline," we use GraphLab to solve a novel three-dimensional retinal image denoising task. In this task we begin with raw three-dimensional laser density estimates, then use GraphLab to generate composite statistics, learn parameters for a large three-dimensional grid pairwise MRF, and then finally compute expectations for each voxel using Loopy BP. Each of these tasks requires both

---

**Algorithm 2**: BP update function

BPUpdate($D_v, D_{*\to v}, D_{v\to *} \in \mathcal{S}_v$) **begin**
  Compute the local belief $b(x_v)$ using $\{D_{*\to v} D_v\}$
  **foreach** $(v \to t) \in (v \to *)$ **do**
    Update $m_{v\to t}(x_t)$ using $\{D_{*\to v}, D_v\}$ and $\lambda_{\text{axis}(vt)}$ from the SDT.
    residual $\leftarrow \left\lVert m_{v\to t}(x_t) - m^{\text{old}}_{v\to t}(x_t) \right\rVert_1$
    **if** *residual > Termination Bound* **then**
      AddTask($t$, residual)
    **end**
  **end**
**end**

---

**Algorithm 3**: Parameter Learning Sync

Fold(acc, vertex) **begin**
  Return acc + image statistics on vertex
**end**
Apply(acc) **begin**
  Apply gradient step to $\lambda$ using acc and return $\lambda$
**end**

---

local iterative computation and global aggregation as well as several different computation schedules.

We begin by using the GraphLab data-graph to build a large (256x64x64) three dimensional MRF in which each vertex corresponds to a voxel in the original image. We connect neighboring voxels in the 6 axis aligned directions. We store the density observations and beliefs in the vertex data and the BP messages in the directed edge data. As shared data we store three global edge parameters which determine the smoothing (accomplished using a Laplace similarity potentials) in each dimension. Prior to learning the model parameters, we first use the GraphLab sync mechanism to compute axis-aligned averages as a proxy for "ground-truth" smoothed images along each dimension. We then performed simultaneous learning and inference in GraphLab by using the background sync mechanism (Alg. 3) to aggregate inferred model statistics and apply a gradient descent procedure. To the best of our knowledge, this is the first time graphical model parameter learning and BP inference have been done concurrently.

**Results:** In Fig. 4(a) we plot the speedup of the parameter learning algorithm, executing inference and learning sequentially. We found that the Splash scheduler outperforms other scheduling techniques enabling a factor 15 speedup on 16 cores. We then evaluated simultaneous parameter learning and inference by allowing the sync mechanism to run concurrently with inference (Fig. 4(b) and Fig. 4(c)). By running a background sync at the right frequency, we found that we can further accelerate parameter learning while only marginally affecting the learned parameters. In Fig. 4(d) and Fig. 4(e) we plot examples of noisy and denoised cross sections respectively.

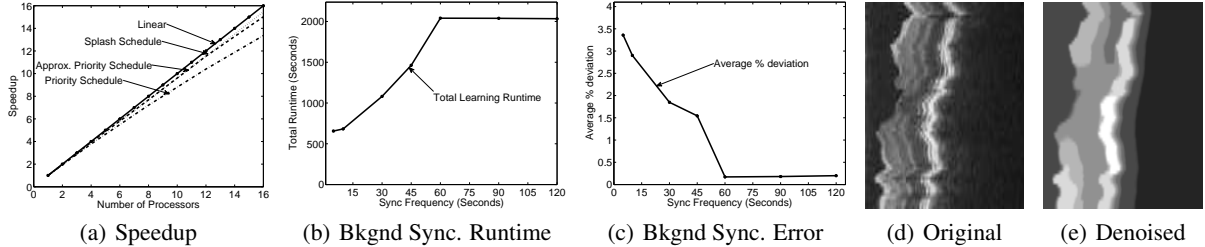

(a) Speedup  (b) Bkgnd Sync. Runtime  (c) Bkgnd Sync. Error  (d) Original  (e) Denoised

Figure 4: *Retinal Scan Denoising* **(a)** Speedup relative to the best single processor runtime of parameter learning using priority, approx priority, and Splash schedules. **(b)** The total runtime in seconds of parameter learning and **(c)** the average percent deviation in learned parameters plotted against the time between gradient steps using the Splash schedule on 16 processors. **(d,e)** A slice of the original noisy image and the corresponding expected pixel values after parameter learning and denoising.

## 4.2 GIBBS SAMPLING

The Gibbs sampling algorithm is inherently sequential and has frustrated efforts to build asymptotically consistent parallel samplers. However, a standard result in parallel algorithms [Bertsekas and Tsitsiklis, 1989] is that for any fixed length Gauss-Seidel schedule there exists an equivalent parallel execution which can be derived from a coloring of the dependency graph. We can extract this form of parallelism using the GraphLab framework. We first use GraphLab to construct a greedy graph coloring on the MRF and then to execute an exact parallel Gibbs sampler.

We implement the standard greedy graph coloring algorithm in GraphLab by writing an update function which examines the colors of the neighboring vertices of $v$, and sets $v$ to the first unused color. We use the edge consistency model with the parallel coloring algorithm to ensure that the parallel execution retains the same guarantees as the sequential version. The parallel Gauss-Seidel schedule is then built using the GraphLab set scheduler (Sec. 3.4.1) and the coloring of the MRF. The resulting schedule consists of a sequence of vertex sets $S_1$ to $S_C$ such that $S_i$ contains all the vertices with color $i$. The *vertex consistency* model is sufficient since the coloring ensures full sequential consistency.

To evaluate the GraphLab parallel Gibbs sampler we consider the challenging task of marginal estimation on a factor graph representing a protein-protein interaction network obtained from Elidan et al. [2006] by generating $10,000$ samples. The resulting MRF has roughly $100K$ edges and $14K$ vertices. As a baseline for comparison we also ran a GraphLab version of the highly optimized Splash Loopy BP [Gonzalez et al., 2009b] algorithm.

**Results:** In Fig. 5 we present the speedup and efficiency results for Gibbs sampling and Loopy BP. Using the set schedule in conjunction with the planning optimization enables the Gibbs sampler to achieve a factor of 10 speedup on 16 processors. The execution plan takes 0.05 seconds to compute, an immaterial fraction of the 16 processor running time. Because of the structure of the MRF, a large number of colors (20) is needed and the vertex distribution over colors is heavily skewed. Consequently there is a strong sequential component to running the Gibbs sampler on this model. In contrast the Loopy BP speedup demonstrates considerably better scaling with factor of 15 speedup on 16 processor. The larger cost per BP update in conjunction with the ability to run a fully asynchronous schedule enables Loopy BP to achieve relatively uniform update efficiency compared to Gibbs sampling.

## 4.3 CO-EM

To illustrate how GraphLab scales in settings with large structured models we designed and implemented a parallel version of Co-EM [Jones, Nigam and Ghani, 2000], a semi-supervised learning algorithm for named entity recognition (NER). Given a list of noun phrases (NP) (e.g., "big apple"), contexts (CT) (e.g., "citizen of _"), and co-occurence counts for each NP-CT pair in a training corpus, CoEM tries to estimate the probability (belief) that each entity (NP or CT) belongs to a particular class (e.g., "country" or "person"). The CoEM update function is relatively fast, requiring only a few floating operations, and therefore stresses the GraphLab implementation by requiring GraphLab to manage massive amounts of fine-grained parallelism.

The GraphLab graph for the CoEM algorithm is a bipartite graph with each NP and CT represented as a vertex, connected by edges with weights corresponding to the co-occurence counts. Each vertex stores the current estimate of the belief for the corresponding entity. The update function for CoEM recomputes the local belief by taking a weighted average of the adjacent vertex beliefs. The adjacent vertices are rescheduled if the belief changes by more than some predefined threshold ($10^{-5}$).

We experimented with the following two NER datasets obtained from web-crawling data.

| Name | Classes | Verts. | Edges | 1 CPU Runtime |
|------|---------|--------|-------|---------------|
| small | 1 | 0.2 mil. | 20 mil. | 40 min |
| large | 135 | 2 mil. | 200 mil. | 8 hours |

We plot in Fig. 6(a) and Fig. 6(b) the speedup obtained by the Partitioned Scheduler and the MultiQueue FIFO scheduler, on both small and large datasets respectively. We observe that both schedulers perform similarly and achieve nearly linear scaling. In addition, both schedulers obtain similar belief estimates suggesting that the update schedule may not affect convergence in this application.

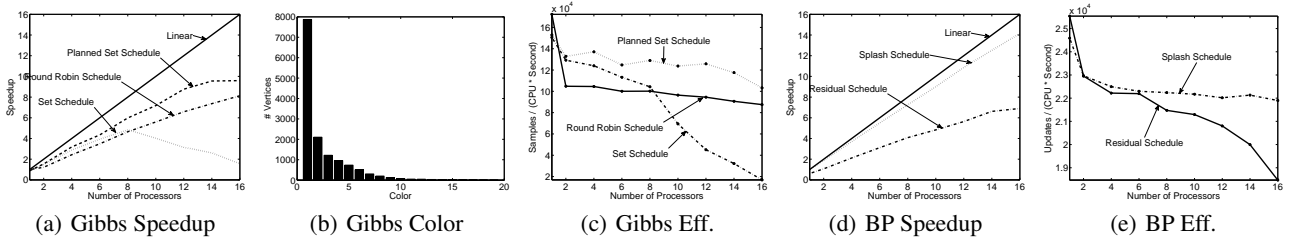

(a) Gibbs Speedup  (b) Gibbs Color  (c) Gibbs Eff.  (d) BP Speedup  (e) BP Eff.

Figure 5: *MRF Inference* **(a)** The speedup of the Gibbs sampler using three different schedules. The *planned set schedule* enables processors to safely execute more than one color simultaneously. The *round robin schedule* executes updates in a fixed order and relies on the edge consistency model to maintain sequential consistency. The plan *set scheduler* does not apply optimization and therefore suffers from substantial synchronization overhead. **(b)** The distribution of vertices over the 20 colors is strongly skewed resulting in a high sequential set schedule. **(c)** The sampling rate per processor plotted against the number of processor provides measure of parallel overhead which is substantially reduced by the plan optimization in the set scheduler. **(d)** The speedup for Loopy BP is improved substantially by the Splash. **(e)** The efficiency of the GraphLab framework as function of the number of processors.

With 16 parallel processors, we could complete three full Round-robin iterations on the large dataset in less than 30 minutes. As a comparison, a comparable Hadoop implementation took approximately 7.5 hours to complete the exact same task, executing on an average of 95 cpus. [Personal communication with Justin Betteridge and Tom Mitchell, Mar 12, 2010]. Our large performance gain can be attributed to data persistence in the GraphLab framework. Data persistence allows us to avoid the extensive data copying and synchronization required by the Hadoop implementation of MapReduce.

Using the flexibility of the GraphLab framework we were able to study the benefits of dynamic (Multiqueue FIFO) scheduling versus a regular round-robin scheduling in CoEM. Fig. 6(c) compares the number of updates required by both schedules to obtain a result of comparable quality on the larger dataset. Here we measure quality by $L_1$ parameter distance to an empirical estimate of the fixed point $x^*$, obtained by running a large number of synchronous iterations. For this application we do not find a substantial benefit from dynamic scheduling.

We also investigated how GraphLab scales with problem size. Figure 6(d) shows the maximum speedup on 16 cpus attained with varying graph sizes, generated by sub-sampling a fraction of vertices from the large dataset. We find that parallel scaling improves with problem size and that even on smaller problems GraphLab is still able to achieve a factor of 12 speedup on 16 cores.

### 4.4 LASSO

The Lasso [Tibshirani, 1996] is a popular feature selection and shrinkage method for linear regression which minimizes the objective $L(w) = \sum_{j=1}^{n}(w^T x_j - y_j)^2 + \lambda \, ||w||_1$.

Unfortunately, there does not exist, to the best of our knowledge, a parallel algorithm for fitting a Lasso model. In this section we implement 2 different parallel algorithms for solving the Lasso.

#### 4.4.1 Shooting Algorithm

We use GraphLab to implement the Shooting Algorithm [Fu, 1998], a popular Lasso solver, and demonstrate that GraphLab is able to *automatically* obtain parallelism by identifying operations that can execute concurrently while retaining sequential consistency.

The shooting algorithm works by iteratively minimizing the objective with respect to each dimension in $w$, corresponding to coordinate descent. We can formulate the Shooting Algorithm in the GraphLab framework as a bipartite graph with a vertex for each weight $w_i$ and a vertex for each observation $y_j$. An edge is created between $w_i$ and $y_j$ with weight $X_{i,j}$ if and only if $X_{i,j}$ is non-zero. We also define an update function (Alg. 4) which operates only on the weight vertices, and corresponds exactly to a single minimization step in the shooting algorithm. A round-robin scheduling of Alg. 4 on all weight vertices corresponds exactly to the sequential shooting algorithm. We automatically obtain an equivalent parallel algorithm by select the full consistency model. Hence, by encoding the shooting algorithm in GraphLab we are able to discover a sequentially consistent *automatic parallelization*.

We evaluate the performance of the GraphLab implementation on a financial data set obtained from Kogan et al. [2009]. The task is to use word counts of a financial report to predict stock volatility of the issuing company for the consequent 12 months. Data set consists of word counts for 30K reports with the related stock volatility metrics.

To demonstrate the scaling properties of the full consistency model, we create two datasets by deleting common words. The sparser dataset contains 209K features and 1.2M non-zero entries, and the denser dataset contains 217K features and 3.5M non-zero entries. The speedup

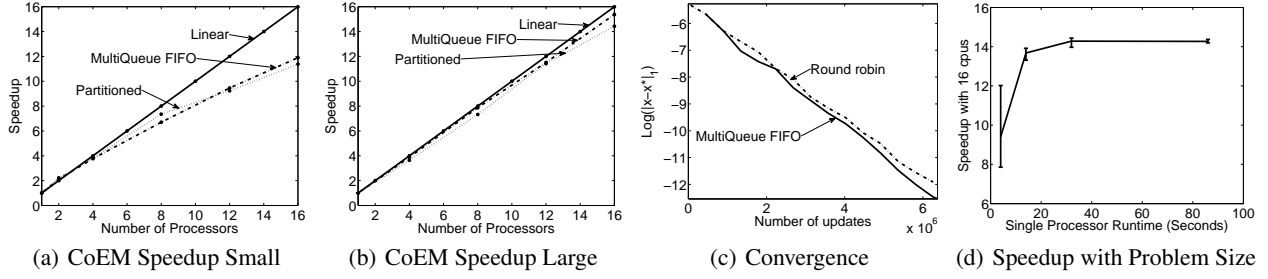

(a) CoEM Speedup Small  (b) CoEM Speedup Large  (c) Convergence  (d) Speedup with Problem Size

Figure 6: *CoEM Results* **(a,b)** Speedup of MultiQueue FIFO and Partitioned Scheduler on both datasets. Speedup is measured relative to fastest running time on a single cpu. The large dataset achieves better scaling because the update function is slower. **(c)** Speed of convergence measured in number of updates for MultiQueue FIFO and Round Robin (equivalent to synchronized Jacobi schedule), **(d)** Speedup achieved with 16 cpus as the graph size is varied.

---

**Algorithm 4**: Shooting Algorithm

ShootingUpdate($D_{w_i}, D_{*\to w_i}, D_{w_i \to *}$) **begin**
    Minimize the loss function with respect to $w_i$
    **if** $w_i$ *changed by* $> \epsilon$ **then**
        Revise the residuals on all $y's$ connected to $w_i$
        Schedule all $w's$ connected to neighboring $y's$
    **end**
**end**

---

**Algorithm 5**: Compressed Sensing Outer Loop

**while** $duality\_gap \geq \epsilon$ **do**
    Update edge and node data of the data graph.
    Use GraphLab to run GaBP on the graph
    Use Sync to compute duality gap
    Take a newton step
**end**

### 4.5 Compressed Sensing

To show how GraphLab can be used as a subcomponent of a larger *sequential* algorithm, we implement a variation of the interior point algorithm proposed by Kim et al. [2007] for the purposes of compressed sensing. The aim is to use a sparse linear combination of basis functions to represent the image, while minimizing the reconstruction error. Sparsity is achieved through the use of elastic net regularization.

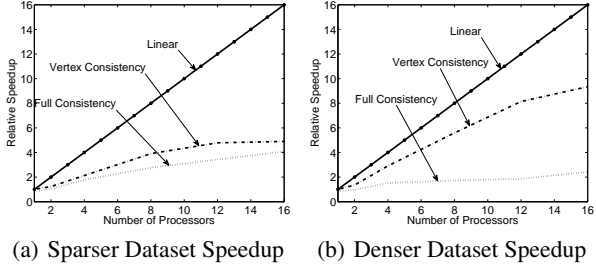

(a) Sparser Dataset Speedup   (b) Denser Dataset Speedup

Figure 7: *Shooting Algorithm* **(a)** Speedup on the sparser dataset using *vertex consistency* and *full consistency* relative to the fastest single processor runtime. **(b)** Speedup on the denser dataset using *vertex consistency* and *full consistency* relative to the fastest single processor runtime.

curves are plotted in Fig. 7. We observed better scaling (4x at 16 CPUs) on the sparser dataset than on the denser dataset (2x at 16 CPUs). This demonstrates that ensuring full consistency on denser graphs inevitably increases contention resulting in reduced performance.

Additionally, we experimented with relaxing the consistency model, and we discovered that the shooting algorithm still converges under the weakest vertex consistency guarantees; obtaining solutions with only $0.5\%$ higher loss on the same termination criterion. The vertex consistent model is much more parallel and we can achieve significantly better speedup, especially on the denser dataset. It remains an open question why the Shooting algorithm still functions under such weak guarantees.

The interior point method is a double loop algorithm where the sequential outer loop (Alg. 5) implements a Newton method while the inner loop computes the Newton step by solving a sparse linear system using GraphLab. We used Gaussian BP (GaBP) as a linear solver [Bickson, 2008] since it has a natural GraphLab representation. The GaBP GraphLab construction follows closely the BP example in Sec. 4.1, but represents potentials and messages analytically as Gaussian distributions. In addition, the outer loop uses a Sync operation on the data graph to compute the duality gap and to terminate the algorithm when the gap falls below a predefined threshold. Because the graph structure is fixed across iterations, we can leverage data persistency in GraphLab, avoid both costly set up and tear down operations and resume from the converged state of the previous iteration.

We evaluate the performance of this algorithm on a synthetic compressed sensing dataset constructed by applying a random projection matrix to a wavelet transform of a $256 \times 256$ Lenna image (Fig. 8). Experimentally, we achieved a factor of 8 speedup using 16 processors using the round-robin scheduling.

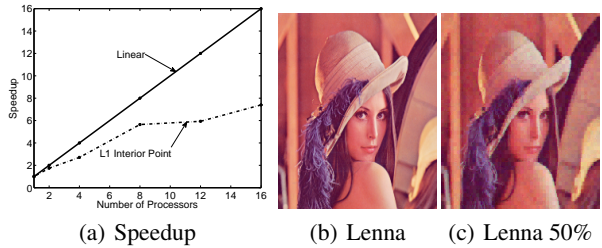

(a) Speedup  (b) Lenna  (c) Lenna 50%

Figure 8: **(a)** Speedup of the Interior Point algorithm on the compressed sensing dataset, **(b)** Original 256x256 test image with 65,536 pixels, **(c)** Output of compressed sensing algorithm using 32,768 random projections.

## 5 CONCLUSIONS AND FUTURE WORK

We identified several limitations in applying existing parallel abstractions like MapReduce to Machine Learning (ML) problems. By targeting common patterns in ML, we developed **GraphLab**, a new parallel abstraction which achieves a high level of usability, expressiveness and performance. Unlike existing parallel abstractions, GraphLab supports the representation of structured data dependencies, iterative computation, and flexible scheduling.

The **GraphLab** abstraction uses a **data graph** to encode the computational structure and data dependencies of the problem. GraphLab represents local computation in the form of **update functions** which transform the data on the **data graph**. Because update functions can modify overlapping state, the GraphLab framework provides a set of data **consistency models** which enable the user to specify the minimal consistency requirements of their application without having to build their own complex locking protocols. To manage sharing and aggregation of global state, GraphLab provides a powerful **sync mechanism**.

To manage the scheduling of dynamic iterative parallel computation, the GraphLab abstraction provides a rich collection of parallel **schedulers** encompassing a wide range of ML algorithms. GraphLab also provides a scheduler construction framework built around a sequence of vertex sets which can be used to compose custom schedules.

We developed an optimized shared memory implementation GraphLab and we demonstrated its performance and flexibility through a series of case studies. In each case study we designed and implemented a popular ML algorithm and applied it to a large real-world dataset achieving state-of-the-art performance.

Our ongoing research includes extending the GraphLab framework to the distributed setting allowing for computation on even larger datasets. While we believe GraphLab naturally extend to the distributed setting we face numerous new challenges including efficient graph partitioning, load balancing, distributed locking, and fault tolerance.

**Acknowledgements**

We thank Guy Blelloch and David O'Hallaron for their guidance designing and implementing GraphLab. This work is supported by ONR Young Investigator Program grant N00014-08-1-0752, the ARO under MURI W911NF0810242, DARPA IPTO FA8750-09-1-0141, and the NSF under grants IIS-0803333 and NeTS-NBD CNS-0721591. Joseph Gonzalez is supported by the AT&T Labs Fellowship Program.